\documentclass{article}
\usepackage{microtype}
\usepackage{graphicx}
\usepackage{subcaption}
\usepackage{booktabs} 

\usepackage{tabularx,booktabs,ragged2e,array}

\usepackage{xcolor}
\usepackage{tcolorbox}
\usepackage{algorithm}
\usepackage{algorithmic}

\usepackage{amsmath}
\usepackage{amssymb}
\usepackage{mathtools}
\usepackage{amsthm}
\usepackage{amsmath,amssymb,amsthm}
\usepackage{enumitem}

\newtheorem{definition}{Definition}
\usepackage[preprint]{neurips_2026}

\usepackage[utf8]{inputenc} 
\usepackage[T1]{fontenc}    
\usepackage{url}            
\usepackage{booktabs}       
\usepackage{amsfonts}       
\usepackage{nicefrac}       
\usepackage{microtype}      
\usepackage{xcolor}         

\title{PTP: Previous-Token Prediction based LLM Inversion\\ for\\ Near-Exact Prompt Reconstruction}

\author{%
  Pirzada Suhail\thanks{Corresponding Author Email: psuhail@iitb.ac.in | Work done during internship at Adobe Research. }\\
  IIT Bombay\\
  \And
  Nagasai Saketh Naidu \\
  Adobe Research \\
  \AND
  Atanu R Sinha \\
  Adobe Research \\
  \And
  Amit Sethi \\
  IIT Bombay \\
}

\begin{document}

\maketitle

\begin{abstract}
Large language models (LLMs) generate text by auto-regressively sampling the next token. This inherently leads to a many-to-many mapping between prompts and responses, complicating the task of inferring prompts from observed outputs. Prior work on LLM inversion frames prompt recovery as a semantic reconstruction task. They rely on fine-tuning pretrained sequence-to-sequence models on large external datasets—and requiring access to model weights or logits—to generate semantically plausible prompts. In contrast, we present a functional approach to inverting a given LLM in a black-box setting, without auxiliary aids. We train an explicit inverse language model entirely from scratch on data synthetically generated from the target LLM itself. Analogous to forward next-token prediction, our inverse model is trained using previous-token prediction, establishing a generative link between the forward and inverse processes that enables faithful prompt reconstruction. Moreover, it naturally supports diverse prompt reconstructions through sampling, whereby all such prompts induce similar responses under the forward, target LLM. Our approach generalises across datasets and exhibits transferability in reconstructing prompts from responses generated by different LLMs. Further, across the set of token based evaluation metrics for prompt and response reconstructions, our approach outperforms prior work.
\end{abstract}
\begin{figure}[h]
\centering
\includegraphics[width=0.55\linewidth]{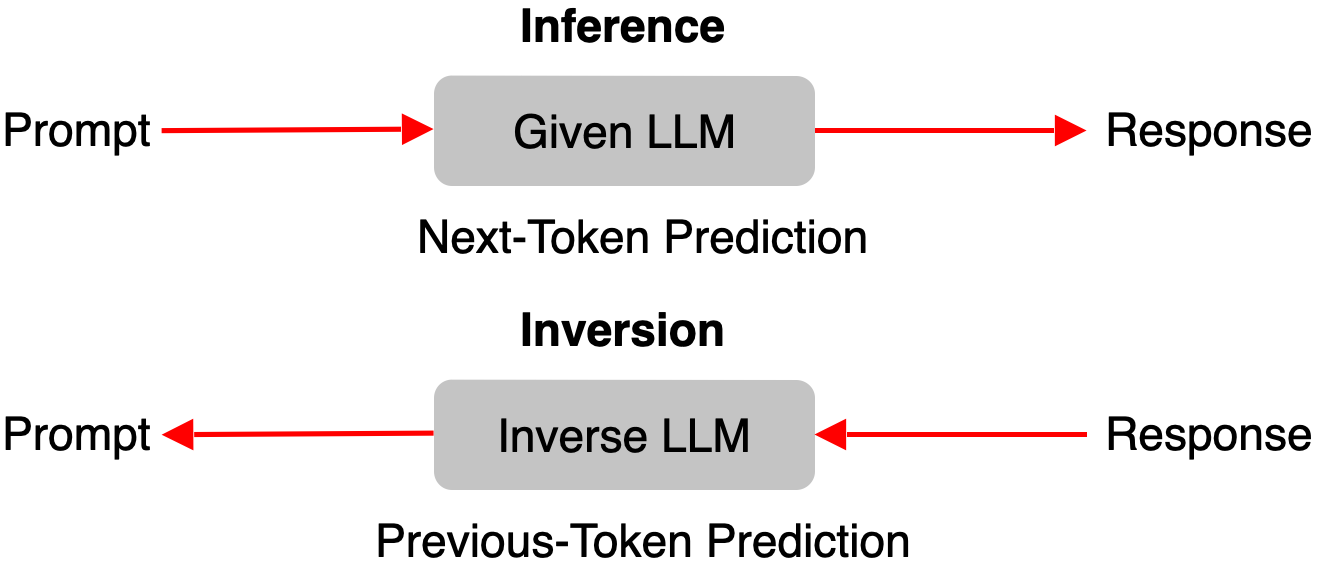}
\caption{Our proposed approach to LLM Inversion using Previous-Token Prediction. 
}
\label{fig:pipeline}
\end{figure}
\section{Introduction}

Large Language Models (LLMs) are typically trained in a forward generative manner using the next-token prediction (NTP) objective. Given a prompt \(x\), a causal language model \( f \) learns to model the conditional probability of the next token at each position. As a result, once trained, the model can generate a sequence auto-regressively from left to right, conditioned on the prompt \( x \). In this work, we aim to \textit{invert} the behavior of a pretrained, black-box LLM entirley from its textual responses. Specifically, given a response \( y \), our goal is to recover a prompt \( \hat{x} \) such that:
\(
f(\hat{x}) \approx y.
\) 

Rather than framing inversion as a semantic reconstruction problem, we propose a functional approach that mirrors the generative structure of the forward model itself, but by going backwards. We train an explicit inverse language model entirely from scratch on data synthetically generated from the target LLM, without access to model weights, logits, or internal representations.

To formalize, consider that at the core of a language model lies a next-token predictor that maps input text to a probability distribution over a fixed vocabulary \( \mathcal{V} \). Specifically, the forward model \(f\) operates as:
\(
f: \text{Context}_{<t} \rightarrow \mathcal{V}, \quad \text{that is,} \quad f(y_{<t}) = P(y_t \mid y_{<t})
\).
In contrast, our inverse model \( f_{\text{inv}} \) is trained to perform the reverse task of predicting the previous token given the future context:
\(
f_{\text{inv}}: \text{Context}_{\geq t} \rightarrow \mathcal{V}, \quad \text{that is,} \quad f_{\text{inv}}(y_{\geq t}) = P(y_{t-1} \mid y_{\geq t})
\)
Thus, while the forward model is trained via NTP, the inverse model learns previous-token prediction (PTP) as shown in Figure~\ref{fig:pipeline}, reconstructing preceding context that would induce the observed response under the given LLM. 

Our general formulation of inversion recovers \( \hat{x} \) that may not be identical to the original prompt which generated \( y \), but results in similar generation. This implies that our approach can discover several prompts \( \hat{x} \) which can yield responses similar to \( y \) which is distinctively useful in controlling undesirable responses through curtailing a diverse set of prompts that may yield those responses. 
\paragraph{Contributions.}
Our main contributions include:
(1) A \emph{previous-token prediction} (PTP) based approach for data-effecient and black-box LLM inversion using an inverse LLM trained from scratch.
(2) A synthetic data generation strategy based on vocabulary-level probing of the target LLM without access to model internals.
(3) The use of PTP-based inversion for faithful prompt reconstruction through greedy reverse decoding and subsequent response reconstruction by re-querying the forward model.
(4) A demonstration of transferability of inverse model by reconstructing prompts from responses generated by different LLMs.

\section{Prior Work}
\paragraph{Network Inversion.}
Network inversion has been studied in \cite{KINDERMANN1990277}  \cite{784232} \cite{SAAD200778}  \cite{Wong2017NeuralNI} as a means of understanding and visualizing the internal mechanisms of neural architectures by recovering input patterns that approximate a given output target.  Beyond interpretability, inversion has also been explored in adversarial settings, with \cite{10.1145/3319535.3354261} \cite{NEURIPS2020_373e4c5d} demonstrating model inversion attacks that reconstruct sensitive features without access to original training data and uncertainty-aware inversion methods \cite{ansari2022autoinverseuncertaintyawareinversion} that constrain inverse solutions to reliable regions of the data distribution. To address optimization challenges in highly non-convex landscapes, \cite{liu2022landscapelearningneuralnetwork} proposed learning a modified loss landscape to accelerate convergence and improve stability. Generative approaches have also emerged, such as \cite{suhail2024networkcnn}, which employs a conditioned generator for inversion of convolutional neural networks while encouraging diversity through feature-space constraints. Finally, deterministic formulations such as \cite{suhail2024network}, which encode networks into conjunctive normal form and apply SAT solvers and samplers to identify inverse solutions.

\paragraph{LLM Inversion.}
\cite{morris2023text} study \emph{embedding inversion}, formulating recovery as controlled generation that iteratively decodes text whose embeddings re-encode close to a target latent vector. Complementarily, \cite{morris2024language} proposes Logit-to-Text(L2T) for \emph{prompt inversion from next-token distributions}, showing that a model’s probability vector over the vocabulary contains rich information about preceding text. Later, Nazir et al.~\cite{nazir2025betterlanguagemodelinversion} propose \emph{Prompt Inversion from Logprob Sequences} (PILS), exploiting the low-dimensional structure of next-token distributions to compactly represent and aggregate log-probabilities across generation steps, yielding substantial gains in exact recovery. Recently, ~\cite{nikolaou2025languagemodelsinjectiveinvertible} introduced \emph{SipIt} suggesting that transformer language models are injective, mapping discrete input sequences to continuous hidden representations enabling provable reconstruction of exact inputs from hidden activations. While effective, these approaches fundamentally rely on access to predictive distributions or internal representations, placing them outside a fully black-box setting.

In black-box regime, \cite{zhang2024effective} provide a systematic evaluation of prompt extraction attacks across multiple LLMs and prompt sources, demonstrating that even simple text-based attacks can reveal hidden system prompts in deployed models. \cite{li2025reversepromptengineering} propose \emph{reverse prompt engineering}(RPE), a training-free black-box approach that queries the target LLM with its own responses and refines candidate prompts via search. Later, \cite{zhang-etal-2024-extracting} introduced Output-to-Prompt(O2P), which recovers prompts using only textual outputs from normal user queries, without access to logits or adversarial inputs.


\newcolumntype{Y}{>{\RaggedRight\arraybackslash\hspace{0pt}}X}
\newcolumntype{L}[1]{>{\RaggedRight\arraybackslash\hspace{0pt}}p{#1}}
\renewcommand\tabularxcolumn[1]{m{#1}}

\begin{table*}[h]
\centering
\caption{Comparison of Prompt reconstruction techniques including L2T (Logit-to-Text) \cite{morris2024language}, 
O2P (Output-to-Prompt) \cite{zhang-etal-2024-extracting}, with our proposed PTP(Previous-Token Prediction).}
\setlength{\tabcolsep}{3pt}          
\renewcommand{\arraystretch}{1.2}    
\label{tab:comp}
\begin{tabularx}{\linewidth}{L{0.23\linewidth}YYY}
\toprule
\textbf{Dimension} & \textbf{L2T} & \textbf{O2P} & \textbf{PTP (Ours)} \\
\midrule
Access required
  & Full Logits
  & Text Outputs
  & Text Outputs \\
Training Data
  & Instructions-2M (2.33M)
  & Instructions-2M (30K prompts)
  & None/Synthetically Generated \\
Responses required
  & $|V| \times$ the number of bits required for desired precision
  & $\sim$64 outputs per prompt
  & Just One \\
Prompts Reconstructed
  & One
  & One
  & Multiple \\
Transferability
  & Poor across LLaMA-2 sizes
  & Moderate zero-shot across GPT-3.5/4, Mistral, Gemma
  & Zero-Shot across GPT-4o \\
\bottomrule
\end{tabularx}
\end{table*}



\paragraph{Comparison.} Popular prompt inversion approaches for LLMs differ substantially in their assumptions, inversion mechanisms, and practical applicability as shown in Table \ref{tab:comp}. Logit-to-Text (L2T) \cite{morris2024language} relies on access to the model’s full next-token probability distributions and uses pretrained sequence-to-sequence models to map these distributions back to prompts. L2T requires querying the model for a large number of logits per token, proportional to the vocabulary size and depends on extensive external training data.

Output-to-Prompt (O2P) by \cite{zhang-etal-2024-extracting} operates purely on textual outputs in a black-box setting by fine-tuning pretrained sequence-to-sequence models to recover prompts from responses. However, O2P requires multiple responses corresponding to the same hidden prompt to achieve effective reconstruction and primarily succeeds at producing semantically similar prompts rather than exact token-level reconstructions.


In contrast, our Previous-Token Prediction (PTP) framework adopts a fundamentally different inversion strategy. Rather than fine-tuning a pretrained sequence-to-sequence model, we train an explicit decoder-based inverse LLM entirely from scratch using synthetically generated data obtained from the target model itself, without access to logits, gradients, embeddings, or original training data. The inverse model mirrors the forward model’s training objective by predicting previous tokens given future context, establishing a principled generative link between forward and inverse processes. This enables near-exact prompt reconstruction at the level of individual tokens when possible, while naturally supporting the recovery of multiple distinct prompts through sampling. Importantly, PTP operates on a single observed response, produces a reusable inverse model, and demonstrates strong zero-shot transferability across different LLMs, all within a strict black-box setting. Together, these properties distinguish our approach as both more practical and more faithful to the underlying generative process than prior inversion techniques.



\section{Methodology}

Our approach to \emph{LLM inversion} is centered on constructing a functional inverse of a pretrained black-box language model by learning an inverse generative model that mirrors the forward model’s auto-regressive structure in reverse. The inverse model is trained entirely from scratch using synthetic sequences generated by probing the forward model itself, without access to external data, pretrained knowledge, or internal signals such as logits or embeddings.


\paragraph{Notation.}
Let \( \mathcal{V} \) denote the finite vocabulary of the given language model, with cardinality \( |\mathcal{V}| \), and let \( \mathcal{V}^* \) denote the set of all finite-length token sequences over \( \mathcal{V} \).
A prompt is represented as a sequence \( x = (x_1, \dots, x_n) \in \mathcal{V}^* \), and a response as a sequence \( y = (y_1, \dots, y_T) \in \mathcal{V}^* \), where all sequences are defined at the token level using a fixed tokenizer.
We consider a frozen causal language model \( f \) which induces a conditional distribution \( P_f(y \mid x) \) over response sequences given a prompt and factorizes auto-regressively as
\(
P_f(y \mid x) = \prod_{t=1}^{T} P_f(y_t \mid x, y_{<t}).
\)
We further denote the inverse language model as \( f_{\text{inv}} \) that induces an auto-regressive conditional distribution over reversed prompt tokens given reversed response context,
\(
P_{f_{\text{inv}}}(\tilde{x} \mid \tilde{y})
= \prod_{t=1}^{n} P_{f_{\text{inv}}}(\tilde{x}_t \mid \tilde{y}, \tilde{x}_{<t}),
\)
where \( \tilde{y} = \text{Reverse}(y) \) and \( \tilde{x} = \text{Reverse}(x) \). 


\begin{definition}[\textbf{Prompt Inversion}]
Let \( f \) be a causal language model inducing a conditional distribution
\( P_f(y \mid x) \) over response sequences \( y \in \mathcal{V}^* \) given
prompt sequences \( x \in \mathcal{V}^* \). Given an observed response sequence \( y \), the \emph{prompt inversion problem}
consists of recovering a prompt sequence \( \hat{x} \in \mathcal{V}^* \) such that
\( \hat{x} \) belongs to the preimage of \( y \) under \( f \), defined as
\[
\mathcal{X}_y
=
\left\{
x \in \mathcal{V}^*
\;\middle|\;
\mathcal{S}\!\left(P_f(\cdot \mid x), y\right)
\leq \varepsilon
\right\},
\]
where \( \mathcal{S}(\cdot,\cdot) \) is a task-dependent similarity measure (e.g. token overlap, semantic similarity)
between response distributions and target responses, and
\( \varepsilon \geq 0 \) denotes the tolerance of equivalence.
\end{definition}

The overall methodology, proposed in Algorithm \ref{alg:ptp_inversion}, can be decomposed into probing the forward model to synthesize data, sequence reversal, training the inverse model on the reversed data, fine-tuning the inverse model on prompt response pairs, and using the inverse model for inference.
\subsection{Forward Model Probing}
\label{probing}
LLMs internalize the statistical and semantic regularities of the training data within their parameters. However, in a black-box setting, without access to its weights we cannot exploit these for our inversion objective. Hence, we conceptualize the forward LLM as a function \(f\) mapping prompts to responses, and the inverse LLM as an approximate inverse of this function. Under this view, functional inversion must remain unaided, i.e., independent of pretrained inverse models or external data. Since we neither have access to model weights nor can use any real data to train the inverse model, we generate synthetic data using the forward model \( f \). Utility of Synthetic Data is evaluated in Appendix Sec. ~\ref{syndi}. 

Training the inverse model on natural text or with pretrained knowledge would compromise this principle, since the inverse model would predict the prompts for a given response from the dataset itself rather than from the inverse mapping of the forward model. 

\begin{definition}[\textbf{Synthetic Data Generation}]
For each token \( v \in \mathcal{V} \) and a fixed generation length \( L \), 
we construct a synthetic sequence
\(
y^{(v)} = (v, y^{(v)}_2, \dots, y^{(v)}_L),
\)
where subsequent tokens are generated auto-regressively according to
\(
y^{(v)}_t \sim P_f(\cdot \mid y^{(v)}_{<t}),
\quad t = 2, \dots, L,
\)
using stochastic decoding with temperature \( \tau \), top-\(k\), and top-\(p\) filtering. The resulting collection
\(
\mathcal{D}_{\text{probe}} = \{ y^{(v)} \mid v \in \mathcal{V} \}
\)
defines our synthetic corpus that collectively capture how the model maps initial tokens to most likely continuations under its learned distribution and is used to train the inverse model, 
\end{definition}

This process results in a collection of sequences which collectively capture how the model maps initial tokens to most likely continuations under its learned distribution. Importantly, these sequences can be generated \textit{on the fly} during training, allowing for efficient memory usage and limitless diversity. This ensures that the sequence generated for the same base token \( v_i \) differs across successive epochs, introducing beneficial randomness into the dataset. Such stochastic probing encapsulates how the forward model transforms input tokens into output distributions, making them informative for learning the inverse mapping in \( f_{\text{inv}} \).
\subsection{Individual Sequence Reversal}
\label{reversal}
The inverse model \( f_{\text{inv}} \) is meant to predict preceding tokens conditioned on future context, such as inferring a plausible prompt from a model-generated response or reconstructing a question given its answer. To achieve this 
using the standard left-to-right next-token prediction objective, we transform each synthetically generated sequence by reversing it at the token level. Specifically, given a sequence \( y^{(i)} = (y^{(i)}_1, \dots, y^{(i)}_L) \) generated by the forward model, we construct its reversed counterpart
\(
\tilde{y}^{(i)} = \text{Reverse}(y^{(i)}) = (y^{(i)}_L, y^{(i)}_{L-1}, \dots, y^{(i)}_1).
\)
This token-level reversal enables the inverse model to maintain compatibility with standard auto-regressive training.

\noindent\textbf{Remark 1.}
Reversal is applied after tokenization rather than at the raw text level. Reversing text at the character or word level can produce token sequences that are different under the model’s tokenizer, even when derived from the same text. By first tokenizing the sequence in the standard forward direction and then reversing the resulting tokens, we preserve the original token inventory.

\subsection{Inverse Model Training}
On the reversed token sequences \( \tilde{y}^{(i)} \), we train the inverse model \( f_{\text{inv}} \) using the standard autoregressive next-token prediction objective. For each reversed sequence \( \tilde{y}^{(i)} = (\tilde{y}_1, \tilde{y}_2, \dots, \tilde{y}_L) \), the model is trained to maximize the log-likelihood of each token conditioned on its left context,
\(
\sum_{t=2}^{L} \log P_{f_{\text{inv}}}(\tilde{y}_t \mid \tilde{y}_{1:t-1}).
\) This corresponds to minimizing the negative log-likelihood loss
\[
\mathcal{L}_{\text{NLL}}(\theta_{\text{inv}})
= - \sum_{t=2}^{L} \log P_{f_{\text{inv}}}(\tilde{y}_t \mid \tilde{y}_{1:t-1}; \theta_{\text{inv}}),
\]
where \( \theta_{\text{inv}} \) denotes the learnable parameters of the inverse model. Optimizing this objective encourages \( f_{\text{inv}} \) to accurately predict each token in the reversed sequence given its causal history, thereby approximating the reverse dynamics of the forward generative process. Since the generated sequence length \( L \) is smaller than the maximum context window \( C \) of the model, we train on the full sequence at every position providing dense token-level supervision and maximizing data efficiency. As a result, the inverse model is trained using the same auto-regressive objective as the forward model, but applied to token-reversed sequences, enabling it to implicitly learn a mapping from future context to previous tokens.

\begin{algorithm}[t]
\caption{Training and Inference using Previous-Token Prediction based LLM Inversion}
\label{alg:ptp_inversion}
\begin{algorithmic}
\REQUIRE Forward LLM \( f \), vocabulary \( \mathcal{V} \), Synthetic Sequence length \( L \)
\ENSURE Trained inverse model \( f_{\text{inv}} \)

\STATE \textbf{Synthetic Sequence Generation:}
\FOR{each token \( v \in \mathcal{V} \)}
    \STATE Generate \( y^{(v)} = (v, y_2, \dots, y_L) \) using \( f \) with stochastic decoding
    \STATE Reverse tokens: \( \tilde{y}^{(v)} \leftarrow \text{Reverse}(y^{(v)}) \)
\ENDFOR

\STATE \textbf{Inverse Model Pre-training:}
\STATE Train \( f_{\text{inv}} \) on \( \{\tilde{y}^{(v)}\} \) using autoregressive NLL loss with Adam Optimiser and lr=1e-3.

\STATE \textbf{Inverse Model Fine-Tuning:}
\FOR{prompt \( x \) in a small downstream set}
    \STATE Generate response \( y \leftarrow f(x) \)
    \STATE Form \( s \leftarrow (x, y) \), reverse \( \tilde{s} \leftarrow \text{Reverse}(s) \)
    \STATE Fine-Tune \( f_{\text{inv}} \) on \( \tilde{s} \) using Adam Optimiser and lr=5e-4.
\ENDFOR

\STATE \textbf{Inference:}
\STATE Given response \( y \), compute \( \tilde{y} \leftarrow \text{Reverse}(y) \)
\STATE Generate reversed prompt \( \tilde{x} \sim f_{\text{inv}}(\cdot \mid \tilde{y}) \)
\STATE Output \( \hat{x} \leftarrow \text{Reverse}(\tilde{x}) \)

\end{algorithmic}
\end{algorithm}

\subsection{Inverse Model Fine-Tuning}
Although the inverse model trained on synthetically generated sequences learns to predict preceding tokens given future context, a brief fine-tuning stage is beneficial to better format the reconstructed prompts. Accordingly, after pretraining \( f_{\text{inv}} \) on the synthetic probing corpus, we further fine-tune it on prompt--response pairs generated by the forward model. This step helps align the inverse model more closely with realistic linguistic patterns and semantics observed in natural language.

To construct the fine-tuning corpus, we sample prompts \( x \in \mathcal{X} \) from a downstream dataset. For each prompt \( x = (x_1, \dots, x_n) \), the frozen forward model \( f \) produces a response \( y = (y_1, \dots, y_T) \) via standard autoregressive decoding. We concatenate the prompt and response to form a single sequence
\(
s = (x_1, \dots, x_n, y_1, \dots, y_T),
\)
which is then reversed at the token level to obtain
\(
\tilde{s} = \text{Reverse}(s) = (y_T, \dots, y_1, x_n, \dots, x_1).
\)
The inverse model is fine-tuned by maximizing the log-likelihood of the reversed sequence under standard autoregressive supervision, or equivalently by minimizing the negative log-likelihood
\[
\mathcal{L}_{\text{NLL}}(\theta_{\text{inv}})
= - \sum_{t=2}^{|\tilde{s}|} \log P_{f_{\text{inv}}}(\tilde{s}_t \mid \tilde{s}_{1:t-1}; \theta_{\text{inv}}),
\]
where \( \theta_{\text{inv}} \) denotes the parameters of the inverse model. This encourages \( f_{\text{inv}} \) to model realistic prompt--response transitions induced by the forward model, improving the reconstruction quality.

\subsection{Inverse Model Inference}
At inference time, given an observed response sequence \( y = (y_1, y_2, \dots, y_T) \), our goal is to reconstruct a plausible prompt \( \hat{x} \) such that conditioning the forward model on \( \hat{x} \) yields a response consistent with \( y \). Inference proceeds by applying the trained inverse model \( f_{\text{inv}} \) in a causal reverse decoding manner.

We first tokenize the response sequence \( y \) using the same tokenizer employed during training and reverse the resulting token sequence to obtain
\(
\tilde{y} = (y_T, y_{T-1}, \dots, y_1).
\)
This reversed sequence \( \tilde{y} \) is provided as context to the inverse model \( f_{\text{inv}} \), which autoregressively generates a sequence of reversed prompt tokens
\(
\tilde{x} = (\tilde{x}_1, \tilde{x}_2, \dots, \tilde{x}_n)
\)
according to the learned conditional distribution. At each step \( t \), tokens may be generated either greedily or via stochastic decoding,
\[
\tilde{x}_t \sim P_{f_{\text{inv}}}(\cdot \mid \tilde{y}, \tilde{x}_{<t}),
\]
with standard decoding controls such as temperature, top-\(k\), or top-\(p\) sampling used to trade off fidelity and diversity. Once generation terminates, the reversed prompt tokens are inverted to recover the prompt in the original order,
\[
\hat{x} = \text{Reverse}(\tilde{x}) = (\tilde{x}_n, \tilde{x}_{n-1}, \dots, \tilde{x}_1).
\]
The reconstructed prompt \( \hat{x} \) can then be passed to the frozen forward model \( f \) to verify if it produces the target response \( y \), or a semantically equivalent response. This inference procedure completes the inversion loop and mirrors standard auto-regressive generation in reverse. See Appendix Sec~\ref{sec:observations} for our observations about training.

\section{Experiments}
We validate our proposed inversion approach across multiple LLMs, including Qwen3-0.6B and LLaMA-2 7B, with different model sizes and architectures on both the \emph{base} and \emph{chat} variants to assess the performance of our approach under instruction tuning and alignment fine-tuning. Although all evaluated models are open source, we restrict ourselves to a \emph{fully black-box setting}, interacting with each model solely through textual input–output queries.

\subsection{Synthetic Sequence Generation}
We experiment with generating synthetic sequences of varying lengths ($64$, $128$, and $256$ tokens) followed by token level reversal as explained in Section 4.1 and 4.2. All generations are performed without instructions, relying solely on next-token prediction with top-$k=5$, top-$p=0.95$, and temperature $0.9$. The sequences are generated \textit{on the fly} during training, allowing for efficient memory usage and limitless diversity. This ensures that the sequence generated for the same base token \( v_i \) differs across successive epochs, introducing beneficial randomness into the dataset. An example of the synthetic sequence generated from the forward model and its corresponding reverse is in the box. While the reversal is shown at word level for simplicity, \textit{actual reversal} is done at token level. 

\begin{tcolorbox}[colback=gray!5!white,colframe=black,title=Example of Synthetic and Reversed Sequences]
\textbf{Synthetic Sequence:} Additionally, what are the health benefits of consuming chocolate, and how does it affect the body?

\textbf{Reversed Sequence:} ? body the affect it does how and, chocolate consuming of benefits health the are what, Additionally
\end{tcolorbox}

\subsection{Training the Inverse Model}
We use Qwen3-0.6B, as our inverse model, which is sufficiently expressive and lightweight enough to be trained from scratch, initialized using He~\cite{he2015delving} initialization. We define an epoch as a complete pass over sequences corresponding to all tokens in the vocabulary. We observe that on-the-fly sequence generation leads to consistently better inversion performance, suggesting that increased stochasticity and diversity during training improves the generalization of the inverse model.

\subsection{Fine-tuning Dataset}
This fine-tuning stage uses a small subset of around 25-30 token long prompts sampled from a dataset with responses generated from the given target LLM to be inverted. The inverse model is then trained using PTP to recover the original prompt from the response. 

For fine-tuning, we make use of the \textbf{ShareGPT} or \textbf{Instructions-2M} dataset that is a collection of user conversations with GPT models, capturing diverse prompt styles and real-world usage patterns. For ShareGPT, we select 500 prompts, using 400 for training and reserving 100 for testing.

\section{Results \& Limitations}\label{sec:res}
We present a comprehensive quantitative analysis of our proposed PTP-based LLM inversion framework. A combination of lexical and semantic similarity metrics, including \textbf{Exact Match (EM)}, \textbf{Token F1}, \textbf{BLEU}, \textbf{ROUGE-L}, \textbf{Cosine Similarity (CS)}, and \textbf{BERT F1} defined in Appendix Sec~\ref{sec:eval-metrics} are used for evaluating the prompt reconstruction quality. For qualitative results see Appendix Sec~\ref{sec:qualitative_results}.

\begin{table*}[h]
    \centering
    \caption{Prompt reconstruction results on Qwen Base (QB) and Qwen Chat (QC).}
    \label{tab:base_chat}
    \resizebox{\textwidth}{!}{%
    \begin{tabular}{lcccccc}
        \toprule
        \textbf{Model} & \textbf{Exact} & \textbf{BLEU} & \textbf{ROUGE-L} & \textbf{CS} & \textbf{BERT F1} & \textbf{Token F1} \\
        \midrule
        Qwen-Base & 35.14 ± 4.4 & 40.58 ± 4.1 & 52.73 ± 3.6 & 74.39 ± 2.2 & 70.49 ± 2.3 & 33.11 ± 4.3 \\
        Qwen-Chat & \textbf{64.77 ± 4.54} & \textbf{63.08 ± 4.60} & \textbf{66.96 ± 4.33} &	\textbf{86.13 ± 2.11} & \textbf{80.66 ± 2.7} & \textbf{63.64 ± 4.61}\\
        \bottomrule
    \end{tabular}%
    }
\end{table*}
\subsection{Base Vs. Chat Models}
We begin by validating the effectiveness of our approach on both the \emph{base} and \emph{chat} variants of Qwen3-0.6B to analyse the impact of instruction fine-tuning on prompt invertibility in Table~\ref{tab:base_chat}. Across all evaluation metrics, the chat model consistently yields substantially higher reconstruction accuracy than the base model, with especially large gains in Exact Match and Token F1. This suggests that instruction tuning does not hinder prompt invertibility; rather, it appears to facilitate inversion.

\begin{table*}[h]
    \centering
    \caption{Prompt reconstruction comparison between Qwen3 0.6B Chat and Llama2 7B Chat with Qwen as the inverse model.}
    \label{tab:qwen_lama}
    \resizebox{\textwidth}{!}{%
    \begin{tabular}{lcccccc}
        \toprule
        \textbf{Model} & \textbf{Exact} & \textbf{BLEU} & \textbf{ROUGE-L} & \textbf{CS} & \textbf{BERT F1} & \textbf{Token F1} \\
        \midrule
        Qwen & \textbf{64.77 ± 4.54} & \textbf{63.08 ± 4.60} & \textbf{66.96 ± 4.33} &	\textbf{86.13 ± 2.11} & \textbf{80.66 ± 2.7} & \textbf{63.64 ± 4.61}\\
        Llama & 17 ± 2.4  & 17 ± 2.7 & 33 ± 2.8 & 84 ± 1.0 & 60 ± 2.0 & 12 ± 1.9 \\
        \bottomrule
    \end{tabular}%
    }
\end{table*}

\subsection{Architecture \& Vocabulary Mismatch}
Next, we evaluate prompt reconstruction across models with differing architectures and vocabularies. Specifically, we compare prompt inversion for responses generated by Qwen3-0.6B Chat and LLaMA-2 7B Chat, while fixing the inverse model to be Qwen3-0.6B Chat in both cases as shown in Table~\ref{tab:qwen_lama}. This setting allows us to isolate the effect of architectural and tokenizer mismatch on inversion performance and assess how well the inverse model generalizes when applied to responses produced by a different model family. As expected, inversion performance is substantially higher when both the forward and inverse models belong to the same family, compared to inverting responses generated by LLaMA-2 7B Chat using a Qwen-based inverse model. The degradation is most pronounced on strict token-level metrics such as Exact Match and Token F1, reflecting vocabulary and tokenization mismatches. In contrast, semantic metrics such as cosine similarity and BERT F1 remain comparatively high, indicating that the inverse model is often able to reconstruct semantically aligned prompts even with architectural and vocabulary differences. 
\begin{table*}[h]
    \centering
    \caption{Transferability Check: Prompt-Response reconstruction results for outputs of Chat-GPT4o inverted using an inverse model trained on Qwen3 0.6B Chat.}
    \label{tab:trans}
    \resizebox{\textwidth}{!}{%
    \begin{tabular}{lcccccc}
        \toprule
        \textbf{P/R} & \textbf{Exact} & \textbf{BLEU} & \textbf{ROUGE-L} & \textbf{CS} & \textbf{BERT F1} & \textbf{Token F1} \\
        \midrule
        Prompt & 11.36 ± 2.79 & 6.56 ± 1.36 & 15.65 ± 1.89 &	\textbf{63.01 ± 1.71} & \textbf{82.26 ± 0.06}  & 10.20 ± 2.77\\
        Response & 1.71 ± 0.43 & 5.02 ± 0.8 & 18.56 ± 1.1 & \textbf{81.57 ± 1.2} & \textbf{84.53 ± 0.38} & 0.65 ± 0.4 \\
        \bottomrule
    \end{tabular}%
    }
\end{table*}

\subsection{Transferability across LLMs}
\textit{Can an inverse model trained on one forward LLM generalize to responses generated by a different LLM?} To evaluate this, we test the inverse model trained on Qwen-3 0.6B Chat against responses produced by GPT-4o on the 100 test prompts. The response from GPT-4o is tokenised using Qwen tokeniser and reversed at the token level for the inverse model to reconstruct the prompt. The reconstructed prompt is then fed back into GPT-4o to generate a reconstructed response. Table~\ref{tab:trans} summarizes the transferability results wherein exact token-level reconstruction is substantially degraded in this cross-model setting for both prompt and response reconstruction because of differences in model architectures and tokenizer. Despite this, semantic similarity metrics remain comparatively high, particularly at the response level indicating that the inverse model is able to reconstruct prompts that preserve sufficient information to induce similar responses when re-injected into the forward model, demonstrating transferability across heterogeneous LLMs.

\begin{table*}[h]
    \centering
    \caption{Cross-dataset generalization: Prompt-Response reconstruction results for inverse model fine-tuned on Instruction-2M (30K samples) and evaluated on the ShareGPT test set.}
    \label{tab:cross_dataset}
    \resizebox{\textwidth}{!}{%
    \begin{tabular}{lcccccc}
        \toprule
        \textbf{P/R} & \textbf{Exact} & \textbf{BLEU} & \textbf{ROUGE-L} & \textbf{CS} & \textbf{BERT F1} & \textbf{Token F1} \\
        \midrule
        Prompt & 11.03 ± 2.19 & 25.80 ± 2.59 & 51.84 ± 2.76 & 73.11 ± 2.05 & 91.34 ± 0.57 & 9.13 ± 1.99 \\
        Response & 4.44 ± 0.84 & 13.25 ± 1.42 & 23.82 ± 1.45 & 82.61 ± 1.78 & 85.14 ± 0.44 & 3.44 ± 0.81 \\
        \bottomrule
    \end{tabular}%
    }
\end{table*}

\subsection{Generalization across Datasets}
To evaluate cross-dataset generalization, we train the inverse model on the Instruction-2M dataset (30K samples) using the Qwen3-0.6B forward model and test it on the ShareGPT test set. The results reported in Table~\ref{tab:cross_dataset} highlight that Semantic-level transfer is robust across datasets while token-level exact metrics are modest under cross-dataset transfer. Despite being fine-tuned on a different dataset, the inverse model retains strong semantic-level generalization as cosine similarity and BERT F1 remain high at both prompt and response level. This indicates that the inverse model learns a generalizable semantic mapping from responses to prompts that transfers across distributions.

\subsection{Baseline Comparisons - PTP Vs. O2P Vs. RPE}
We next compare our PTP-based LLM inversion framework against prior SOTA method for prompt reconstruction. Since PTP operates entirely in a fully black-box setting using only textual model outputs, we restrict the comparison to Output-to-Prompt (O2P)~\cite{zhang-etal-2024-extracting} and Reverse Prompt Engineering (RPE)~\cite{li2025reversepromptengineering}, a recent training-free LLM-as-optimizer framework.

Unlike PTP, RPE treats prompt inversion as a search problem and relies on iterative refinement coupled with genetic algorithms to evolve candidate prompts that reproduce the observed response. RPE repeatedly queries a powerful LLM to optimize candidate prompts. In contrast, O2P relies heavily on pretrained sequence-to-sequence models and large auxiliary datasets, which significantly contribute to its performance. To account for this disparity, we compare our method against multiple O2P variants operating under increasingly favorable data conditions. PTP performs single-shot prompt reconstruction, without any iterative refinement, evolutionary search, or population-based optimization. While incorporating iterative refinement or genetic algorithms could plausibly improve PTP performance, doing so would introduce substantial computational overhead and fundamentally change the method’s cost profile. We therefore deliberately restrict PTP to a single-pass reconstruction and evaluate RPE in its strongest reported configuration. 

O2P and RPE evaluates inversion solely based on the ability to reconstruct the original prompt. In contrast, we additionally assess whether reconstructed prompts preserve the forward model’s behavior. Specifically, after reconstructing prompts, we re-query the forward model to generate responses and compare these \emph{reconstructed responses} against the original responses. This closed-loop evaluation reveals whether diverse reconstructed prompts still induce semantically similar model behavior. We evaluate all approaches on inversion of Qwen3-0.6B Chat to ensure a fair comparison on 100 test samples drawn from the ShareGPT dataset~\cite{zhang-etal-2024-extracting} in following configurations:
\begin{enumerate}[leftmargin=*, itemsep=0pt, topsep=0pt, parsep=0pt]
    \item \textbf{PTP (Ours)}: Pre-training Qwen from scratch with our inversion method, followed by fine-tuning on 400 prompts(only) from the ShareGPT dataset.
    
    \item \textbf{O2P-1}: T5 is pretrained using outputs from Qwen with Instruction 2M--30k samples dataset, as in ~\cite{zhang-etal-2024-extracting}, followed by finetuning on 400 ShareGPT samples.
    
    \item \textbf{O2P-2}: Pre-trained T5 is fine-tuned on Instruction 2M--30k samples dataset with no further fine-tuning on ShareGPT dataset.
    
    \item \textbf{O2P-3}: The pre-trained base T5 model is trained on 400 samples from the ShareGPT dataset without exposure to the larger Instruction 2M--30k samples dataset.
    
    \item \textbf{RPEGA}: The best-performing variant from~\cite{li2025reversepromptengineering} leveraging iterative evolutionary optimization evaluated on the same 100-sample ShareGPT test set.
\end{enumerate}
\begin{table*}[t]
    \centering
    \caption{Prompt reconstruction results on Qwen3 0.6B Chat using PTP (Ours), O2P, and RPE.}
    \label{tab:prompt_recon}
    \resizebox{\textwidth}{!}{%
    \begin{tabular}{lcccccc}
        \toprule
        \textbf{Model} & \textbf{Exact} & \textbf{BLEU} & \textbf{ROUGE-L} & \textbf{CS} & \textbf{BERT F1} & \textbf{Token F1} \\
        \midrule
        PTP (Ours) & \textbf{64.77 ± 4.54} & \textbf{63.08 ± 4.60} & 66.96 ± 4.33 & 86.13 ± 2.11 & 80.66 ± 2.70 & \textbf{63.64 ± 4.61} \\
        O2P-1 & 25.44 ± 3.00 & 40.59 ± 2.61 & \textbf{69.92 ± 1.81} & \textbf{89.36 ± 0.76} & \textbf{84.09 ± 1.00} & 21.00 ± 2.72 \\
        O2P-2 & 19.66 ± 2.49 & 34.28 ± 2.33 & 64.25 ± 1.87 & 87.70 ± 0.74 & 80.83 ± 1.01 & 15.48 ± 2.27 \\
        O2P-3 & 19.86 ± 2.57 & 33.45 ± 2.56 & 64.90 ± 2.02 & 86.97 ± 0.84 & 80.91 ± 1.10 & 16.05 ± 2.35 \\
        RPEGA & 0.64 ± 1.74 & 2.35 ± 1.87 & 11.90 ± 4.32 & 65.29 ± 8.57 & 50.97 ± 4.07 & 0.26 ± 0.73 \\
        \bottomrule
    \end{tabular}%
    }
\end{table*}

\begin{table*}[t]
    \centering
    \caption{Response reconstruction results on Qwen3 0.6B Chat using PTP (Ours), O2P, and RPE.}
    \label{tab:response_recon}
    \resizebox{\textwidth}{!}{%
    \begin{tabular}{lcccccc}
        \toprule
        \textbf{Model} & \textbf{Exact} & \textbf{BLEU} & \textbf{ROUGE-L} & \textbf{CS} & \textbf{BERT F1} & \textbf{Token F1} \\
        \midrule
        PTP (Ours) & \textbf{61.79 ± 4.60} & \textbf{64.87 ± 4.30} & \textbf{71.43 ± 3.60} & 88.08 ± 1.90 & \textbf{85.48 ± 1.90} & \textbf{61.07 ± 4.70} \\
        O2P-1 & 24.17 ± 2.96 & 33.57 ± 2.73 & 64.20 ± 2.00 & 86.16 ± 1.02 & 80.19 ± 1.24 & 19.28 ± 2.74 \\
        O2P-2 & 17.70 ± 2.38 & 27.85 ± 2.45 & 57.19 ± 2.11 & 83.77 ± 1.22 & 76.49 ± 1.27 & 13.09 ± 2.05 \\
        O2P-3 & 17.15 ± 2.43 & 24.83 ± 2.69 & 46.97 ± 3.01 & 70.57 ± 2.69 & 67.66 ± 2.08 & 12.86 ± 2.15 \\
        RPEGA & 3.34 ± 8.44 & 14.39 ± 9.01 & 34.39 ± 11.08 & \textbf{88.81 ± 9.40} & 73.68 ± 5.55 & 1.52 ± 4.43 \\
        \bottomrule
    \end{tabular}%
    }
\end{table*}
\textbf{Prompt Reconstruction:} Table~\ref{tab:prompt_recon} reports prompt reconstruction performance in which our approach substantially outperforms O2P across all token-overlap–based metrics. While RPEGA performs substantially below PTP across all evaluation metrics. The gains are particularly pronounced for exact reconstruction attributed to the direct generative link preserved between the forward and the inverse model via previous-token prediction. While O2P attains high scores on semantic similarity metrics such as cosine similarity and BERT F1, our method achieves comparable performance on these measures despite being trained solely on 256-token synthetic sequences generated from the target LLM. 

\textbf{Response Reconstruction:} PTP-based approach consistently outperforms O2P in response reconstruction Table~\ref{tab:response_recon}, demonstrating that reversing the auto-regressive process yields prompts that are functionally equivalent to the originals. While RPEGA attains moderate cosine similarity at the response level, it fails almost entirely on token level metrics. This suggests that training-free prompted inversion is insufficient for exact prompt recovery, and that a dedicated inverse model trained via PTP cannot be replicated by simply prompting a capable LLM. See Appendix Sec. ~\ref{sec:pr} for Histograms.

\textbf{Observations:} Across the O2P variants, we observe performance progression from O2P-3 to O2P-1 as more training data is introduced, highlighting the strong dependence of O2P on pretrained models and external datasets to achieve effective prompt reconstruction. In contrast, our PTP-based method achieves superior token-level recovery without relying on pretrained models or real-world data. Moreover, the consistent improvements across both lexical and semantic metrics indicate that our approach does not merely reconstruct semantically plausible prompts, but also preserves fine-grained lexical and structural details of the original inputs, underscoring the effectiveness of framing inversion as a generative reverse modeling problem rather than a data-driven semantic reconstruction task.

\subsection{Cross-Backbone Ablations}
Above experiments compare PTP with a decoder-only against O2P with a T5 inverse model . PTP fundamentally relies on reversed left-to-right autoregressive generation, which naturally requires a decoder-only architecture, whereas O2P is formulated as a sequence-to-sequence mapping and is therefore best served by an encoder–decoder model. To rule out architectural confounds, we conduct cross-backbone ablations in the Appendix Sec. ~\ref{ablations} by implementing O2P with Qwen, and running PTP with a T5 backbone. 

In all cases, performance degrades substantially and remains well below our main PTP with Qwen results. Notably, O2P implemented with the same Qwen backbone as PTP still fails to recover prompts at the token level, demonstrating that PTP’s gains are driven by its inversion objective and synthetic pre-training strategy rather than model capacity or backbone choice. Similarly, implementing PTP using a pretrained T5 backbone leads to substantial degradation across all metrics. This is consistent with the expected directional penalty introduced when forcing T5 to perform reversed auto-regressive generation, which conflicts with the left-to-right sequence modeling priors encoded during pretraining.

\subsection{Limitations and Future Work}
Despite strong performance on semantic reconstruction, exact token level recovery remains sensitive to target LLM tokenizer access, vocabulary and architecture mismatch. While pre-training is data free, the current framework still requires lightweight fine-tuning on a minimal set of prompts to recover prompt formatting. Also synthetic data generation over the entire vocabulary of the target LLM introduces a non-trivial query budget. Future work will explore efficient probing strategies and iterative refinement mechanisms to improve scalability and reconstruction quality. More details related to scalability, data-efficiency, black-box assumptions, and probing strategy are provided in Appendix Sec. ~\ref{appendix:practicality}.

\section{Conclusion}
We present a Previous-Token Prediction (PTP) based approach to LLM inversion using an inverse language model trained entirely from scratch on synthetic data for faithful prompt reconstruction. PTP outperforms prior black-box baselines on lexical metrics and preserves semantic behavior under closed-loop response reconstruction. PTP also exhibits meaningful semantic transferability across tokenizers, datasets, and models, when exact lexical prompt recovery is impeded.

\bibliography{neurips_2026}
\bibliographystyle{plainnat}

\newpage
\appendix

\section{Observations about Training}\label{sec:observations}
In pre-training the inverse model, we observe that longer training sequences substantially improve reconstruction performance. Consequently, we report primary results for models trained on $256$-token sequences. Importantly, the framework can be easily extended to longer contexts, up to the forward model’s maximum context length of $4096$ tokens, suggesting significant room for further improvement. 

Additionally, we find that on-the-fly sequence generation introduces greater diversity in the training data, reducing overfitting to specific contexts and leading to improved inversion quality compared to pre-generated sequences. By resampling responses at each epoch using stochastic decoding strategies (top-$k$, top-$p$, and temperature sampling), the same base token can give rise to multiple distinct contexts. This variability enriches the training distribution and allows the inverse model to learn a more robust mapping from responses to prompts, rather than memorizing fixed continuations. 

Another key observation is the benefit of fine-tuning on real-world prompts after synthetic pre-training. While pre-training on synthetic sequences is sufficient to establish the structural inverse mapping, fine-tuning on curated datasets such as ShareGPT further aligns the inverse model with natural linguistic patterns. This two-stage training procedure yields an inverse model that is both structurally principled and semantically aligned, improving its ability to reconstruct realistic prompts beyond synthetic text. The improvements are especially noticeable in qualitative evaluations (Appendix Sec~\ref{sec:qualitative_results}), where fine-tuned models recover prompts that are nearly indistinguishable from the originals. We also note the role of forward model choice: the Qwen Chat variant consistently yields better inversion results compared to Qwen Base. 

We also explore the \textbf{transferability} of the inverse model. Specifically, an inverse model trained on Qwen Chat is able to successfully reconstruct prompts when the responses come from a different forward model such as ChatGPT-4o. This demonstrates that the learned inverse mapping generalizes beyond the specific forward model used for training, highlighting the portability of the approach across architectures. Transferability underscores the practicality of the framework, showing that a single inverse model can invert multiple forward models without retraining.

\newpage
\section{Description of Evaluation Metrics Used}\label{sec:eval-metrics}
To assess the effectiveness of the inverse model, we employ a range of complementary evaluation metrics that capture both exact lexical similarity and broader semantic alignment between reconstructed and ground-truth prompts:

\begin{itemize}
    \item \textbf{Exact Match (EM):} A strict measure that checks whether the reconstructed prompt is identical to the ground-truth prompt at the token level.
    \item \textbf{BLEU:} A precision-oriented metric based on $n$-gram overlap, sensitive to exact word order and phrasing.
    \item \textbf{ROUGE-L:} A recall-oriented metric that evaluates the longest common subsequence (LCS) between predicted and reference text, capturing partial overlaps and fluency.
    \item \textbf{Cosine Similarity (CS):} Semantic similarity computed between sentence embeddings of the reconstructed and original prompts, using pretrained encoders. This metric is robust to paraphrasing.
    \item \textbf{BERT F1:} A semantic evaluation metric that aligns contextual embeddings from a pretrained BERT model and computes an F1 score over matches, capturing nuanced meaning beyond surface-level overlap.
    \item \textbf{Token F1:} Computes precision, recall, and F1 scores based on token overlap, providing a more flexible measure than exact match.
\end{itemize}

\newpage
\section{Practicality, Data Dependence, and Black-Box Assumptions}
\label{appendix:practicality}

In this section, we address several practical concerns regarding the scalability, data dependence, black-box assumptions, and probing strategy underlying PTP.

\subsection{Efficiency and Practicality}

A key practical concern for black-box inversion methods is the query budget required to construct the synthetic pre-training corpus. We therefore provide a concrete estimate of the computational cost of synthetic data generation used in PTP.

The vocabulary size of Qwen3-0.6B is approximately
\[
|\mathcal{V}| \approx 150{,}000.
\]
For each token \(v \in \mathcal{V}\), we generate a single synthetic sequence of length \(L\) using the forward model. The total number of queried forward-model tokens during one epoch of synthetic pre-training is therefore
\[
|\mathcal{V}| \times L
\;\approx\;
150{,}000 \times 256
\;\approx\;
38\text{M tokens}.
\]

Although this budget scales linearly with the number of pre-training epochs under on-the-fly generation, the process remains computationally practical because each sequence is short and generated independently. Synthetic generation is therefore trivially parallelizable across GPUs and batched efficiently in practice. On a single A100 GPU, one epoch of synthetic corpus generation takes approximately one hour.

\subsection{Data Dependence and the Role of Fine-Tuning}

Our claim of being ``data-free'' pertains specifically to the synthetic pre-training stage. During pre-training, the inverse model is trained entirely on synthetic sequences generated directly from the target LLM without relying on any external instruction-response datasets.

The subsequent fine-tuning stage uses a minimal set of only 400 prompts from ShareGPT, compared to the 30K+ external instruction-response pairs used in O2P's main training pipeline. Critically, this fine-tuning stage serves only as a lightweight adaptation step rather than the source of the inverse capability itself.

Specifically:

\begin{enumerate}[leftmargin=*, label=(\roman*)]
    \item Only the \emph{prompts} are drawn from ShareGPT. All responses used during fine-tuning are generated on the fly using the frozen forward model \(f\).

    \item The inverse model is trained entirely from scratch and does not inherit pretrained inversion capabilities from another model.

    \item Appendix Sec. ~\ref{syndi} demonstrates empirically that synthetic pre-training is essential for learning a meaningful inverse mapping. Models trained without synthetic pre-training exhibit high validation loss and fail to recover prompts reliably from limited real prompts alone. In contrast, models initialized with synthetic pre-training achieve substantially stronger reconstruction performance and validation accuracy.

    \item Appendix~ Sec. ~\ref{syndi} further shows that fine-tuning contributes primarily to formatting refinement rather than learning the inverse mapping itself. Models that undergo synthetic pre-training show minimal changes in validation loss and reconstruction accuracy during the fine-tuning stage, indicating that the inverse relationship is already established prior to fine-tuning.
\end{enumerate}

Taken together, these results indicate that the central inversion capability of PTP arises from synthetic pre-training rather than from dependence on external supervised datasets.

\newpage

\subsection{Semi-Black-Box Assumptions and Tokenizer Access}

We acknowledge that tokenizer access is required for \emph{token-level exact reconstruction}, where the goal is to recover the precise sequence of token IDs corresponding to the original prompt.

However, it is important to distinguish token-level reconstruction from semantic reconstruction. Tokenizer access is not required for semantic prompt recovery. When the objective is semantic reconstruction rather than exact token recovery, the inverse model may operate entirely with its own tokenizer and vocabulary independently of the target model.

Tokenizer access becomes necessary only because our primary contribution focuses on strict token-level exact reconstruction, which requires consistency between the forward and inverse tokenization processes. We therefore view tokenizer access as an explicit assumption tied to the stricter evaluation setting rather than a hidden dependency of the method itself.

\subsection{Sufficiency of Single-Token Vocabulary Probing}

A natural question is whether probing the forward model using only single-token seeds sufficiently captures the structure of the prompt distribution.

The goal of single-token probing is not to reproduce the distribution of natural prompts directly. Instead, the objective is to expose the reverse conditional structure of the forward model across its entire vocabulary.

For every token \(v \in \mathcal{V}\), we generate synthetic continuations conditioned on that token. Collectively, these generated sequences cover the model's token-transition dynamics across the vocabulary and expose the statistical regularities governing how the model maps contexts to continuations.

Moreover, stochastic generation with temperature, top-\(k\), and top-\(p\) sampling ensures that the same seed token generates distinct trajectories across epochs. On-the-fly generation therefore substantially diversifies the synthetic corpus despite the simplicity of the probing seed.

Using larger seed combinations would require introducing assumptions about token ordering and prompt structure that are external to the forward model itself. Single-token probing avoids imposing such assumptions while preserving the data-independent nature of synthetic pre-training.

\newpage
\section{Cross-Backbone Ablations}
\label{ablations}

In the main paper, PTP is implemented using a decoder-only autoregressive inverse model, whereas the O2P baseline uses a T5 encoder--decoder architecture. A natural question is whether the observed gains arise from the inversion objective itself or simply from differences in backbone architecture. 

We argue that direct backbone matching between PTP and O2P is fundamentally non-trivial because the architecture choice is dictated by the methodological formulation of each approach.

First, PTP fundamentally relies on reversed left-to-right autoregressive generation. The core mechanism of PTP is previous-token prediction over reversed sequences, which establishes a generative reverse process linking outputs back to prompts. This formulation naturally requires a decoder-only causal language model capable of strict autoregressive generation. In contrast, T5 is an encoder--decoder architecture that first encodes the entire input sequence before decoding the output sequence, making it fundamentally mismatched with the reversed autoregressive paradigm used in PTP.

Second, the value of T5 within O2P comes primarily from its pretrained sequence-to-sequence priors. Training T5 from scratch merely to match the training regime of PTP would remove the very advantage that motivates its use in O2P, making such a comparison uninformative. Conversely, adapting pretrained T5 to perform PTP requires fine-tuning it on reversed prompt--response sequences, which introduces a strong directional mismatch with the left-to-right generative priors encoded during pretraining. Fine-tuning T5 to autoregressively generate reversed sequences therefore antagonizes the pretrained priors baked into the decoder. Any resulting performance degradation should thus be attributed to degradation of pretrained priors rather than to limitations of the PTP objective itself.

Finally, the backbone choice reflects a deeper methodological distinction between the two approaches. O2P formulates inversion as a sequence-to-sequence mapping problem, naturally suited to encoder--decoder architectures. In contrast, PTP formulates inversion as a reverse autoregressive generation problem, naturally suited to decoder-only causal language models.

Nevertheless, to provide the most complete comparison possible, we conduct cross-backbone ablations in both directions: implementing O2P using the same Qwen3-0.6B decoder-only backbone used by PTP, and implementing PTP using a T5 backbone. Tables~\ref{tab:ablation_prompt} and~\ref{tab:ablation_response} report prompt and response reconstruction results respectively.

\begin{table*}[h]
    \centering
    \caption{Prompt reconstruction results for cross-backbone ablations. O2P is implemented using the same Qwen3-0.6B backbone as PTP, while PTP is additionally evaluated using a T5 inverse model.}
    \label{tab:ablation_prompt}
    \resizebox{\textwidth}{!}{%
    \begin{tabular}{lcccccc}
        \toprule
        \textbf{Method} & \textbf{Exact} & \textbf{BLEU} & \textbf{ROUGE-L} & \textbf{CS} & \textbf{Token F1} & \textbf{BERT F1} \\
        \midrule
        O2P-1 (Qwen) & 12.39 $\pm$ 20.38 & 5.99 $\pm$ 3.65 & 21.46 $\pm$ 5.47 & 76.29 $\pm$ 13.62 & 9.31 $\pm$ 16.86 & 58.09 $\pm$ 5.56 \\
        O2P-2 (Qwen) & 4.06 $\pm$ 8.12 & 4.53 $\pm$ 3.30 & 20.19 $\pm$ 5.45 & 71.34 $\pm$ 10.25 & 2.48 $\pm$ 5.73 & 57.16 $\pm$ 7.01 \\
        O2P-3 (Qwen) & 16.10 $\pm$ 22.27 & 5.47 $\pm$ 3.51 & 20.84 $\pm$ 5.10 & 80.33 $\pm$ 9.77 & 12.24 $\pm$ 19.03 & 59.47 $\pm$ 5.52 \\
        PTP (T5) & 2.82 $\pm$ 7.73 & 2.69 $\pm$ 8.54 & 12.58 $\pm$ 11.60 & 47.69 $\pm$ 12.94 & 1.58 $\pm$ 6.04 & 47.05 $\pm$ 8.86 \\
        PTP (Qwen, Ours) & \textbf{64.77 $\pm$ 4.54} & \textbf{63.08 $\pm$ 4.60} & \textbf{66.96 $\pm$ 4.33} & \textbf{86.13 $\pm$ 2.11} & \textbf{63.64 $\pm$ 4.61} & \textbf{80.66 $\pm$ 2.70} \\
        \bottomrule
    \end{tabular}%
    }
\end{table*}

\begin{table*}[h]
    \centering
    \caption{Response reconstruction results for cross-backbone ablations. O2P is implemented using the same Qwen3-0.6B backbone as PTP, while PTP is additionally evaluated using a T5 inverse model.}
    \label{tab:ablation_response}
    \resizebox{\textwidth}{!}{%
    \begin{tabular}{lcccccc}
        \toprule
        \textbf{Method} & \textbf{Exact} & \textbf{BLEU} & \textbf{ROUGE-L} & \textbf{CS} & \textbf{Token F1} & \textbf{BERT F1} \\
        \midrule
        O2P-1 (Qwen) & 1.26 $\pm$ 1.67 & 3.00 $\pm$ 7.27 & 15.68 $\pm$ 12.54 & 67.67 $\pm$ 18.52 & 0.32 $\pm$ 0.47 & 49.52 $\pm$ 10.94 \\
        O2P-2 (Qwen) & 1.33 $\pm$ 1.28 & 4.67 $\pm$ 8.11 & 20.11 $\pm$ 11.62 & 71.06 $\pm$ 19.15 & 0.33 $\pm$ 0.37 & 52.79 $\pm$ 10.56 \\
        O2P-3 (Qwen) & 1.21 $\pm$ 1.23 & 2.98 $\pm$ 4.76 & 19.30 $\pm$ 10.47 & 73.58 $\pm$ 11.59 & 0.39 $\pm$ 0.52 & 52.81 $\pm$ 8.54 \\
        PTP (T5) & 1.90 $\pm$ 1.24 & 1.90 $\pm$ 4.49 & 15.86 $\pm$ 9.03 & 59.27 $\pm$ 21.26 & 0.69 $\pm$ 1.80 & 51.36 $\pm$ 9.59 \\
        PTP (Qwen, Ours) & \textbf{61.79 $\pm$ 4.60} & \textbf{64.87 $\pm$ 4.30} & \textbf{71.43 $\pm$ 3.60} & \textbf{88.08 $\pm$ 1.90} & \textbf{61.07 $\pm$ 4.70} & \textbf{85.48 $\pm$ 1.90} \\
        \bottomrule
    \end{tabular}%
    }
\end{table*}

The results strongly support the conclusions presented in the main paper. Replacing T5 with Qwen as the inverse model for O2P does not improve performance. All O2P variants implemented using the same Qwen backbone as PTP remain substantially below PTP on all token-level metrics for both prompt and response reconstruction. This demonstrates that the performance gains of PTP are not explained by model capacity or backbone choice alone.

Similarly, implementing PTP using a pretrained T5 backbone leads to substantial degradation across all metrics. This is consistent with the expected directional penalty introduced when forcing T5 to perform reversed autoregressive generation, which conflicts with the left-to-right sequence modeling priors encoded during pretraining.

Taken together, these experiments provide strong evidence that PTP's gains arise from the inversion formulation itself --- namely previous-token prediction over synthetically generated sequences --- together with the synthetic pre-training strategy that establishes a principled generative reverse process between responses and prompts.

\newpage

\section{Synthetic Data is Very Helpful}
\label{syndi}

\begin{figure*}[h]
    \centering
    \begin{subfigure}[t]{0.48\textwidth}
        \centering
        \includegraphics[width=\linewidth]{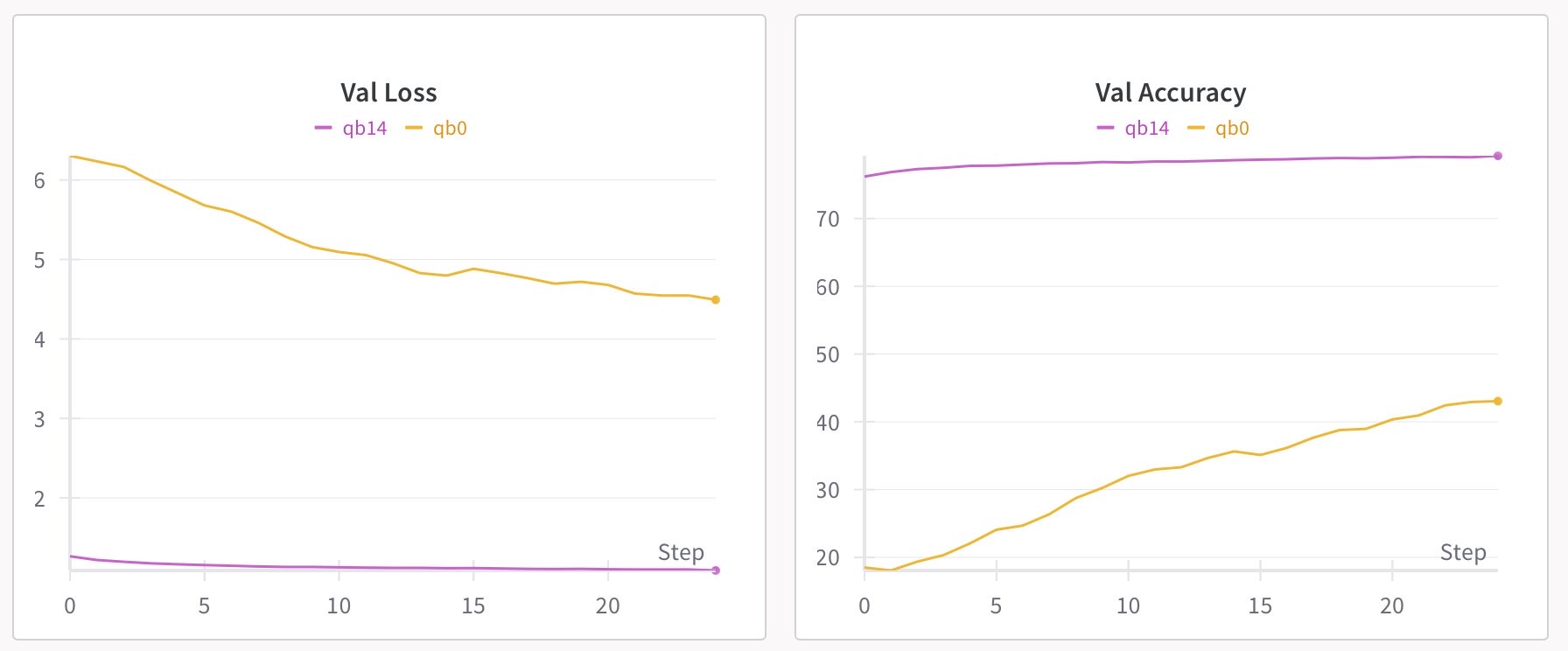}
        \caption{Qwen3-0.6B Base. \textbf{Yellow}: training without synthetic pre-training (ShareGPT only).
        \textbf{Purple}: synthetic pre-training followed by fine-tuning.}
        \label{fig:qwen_base_synth}
    \end{subfigure}
    \hfill
    \begin{subfigure}[t]{0.48\textwidth}
        \centering
        \includegraphics[width=\linewidth]{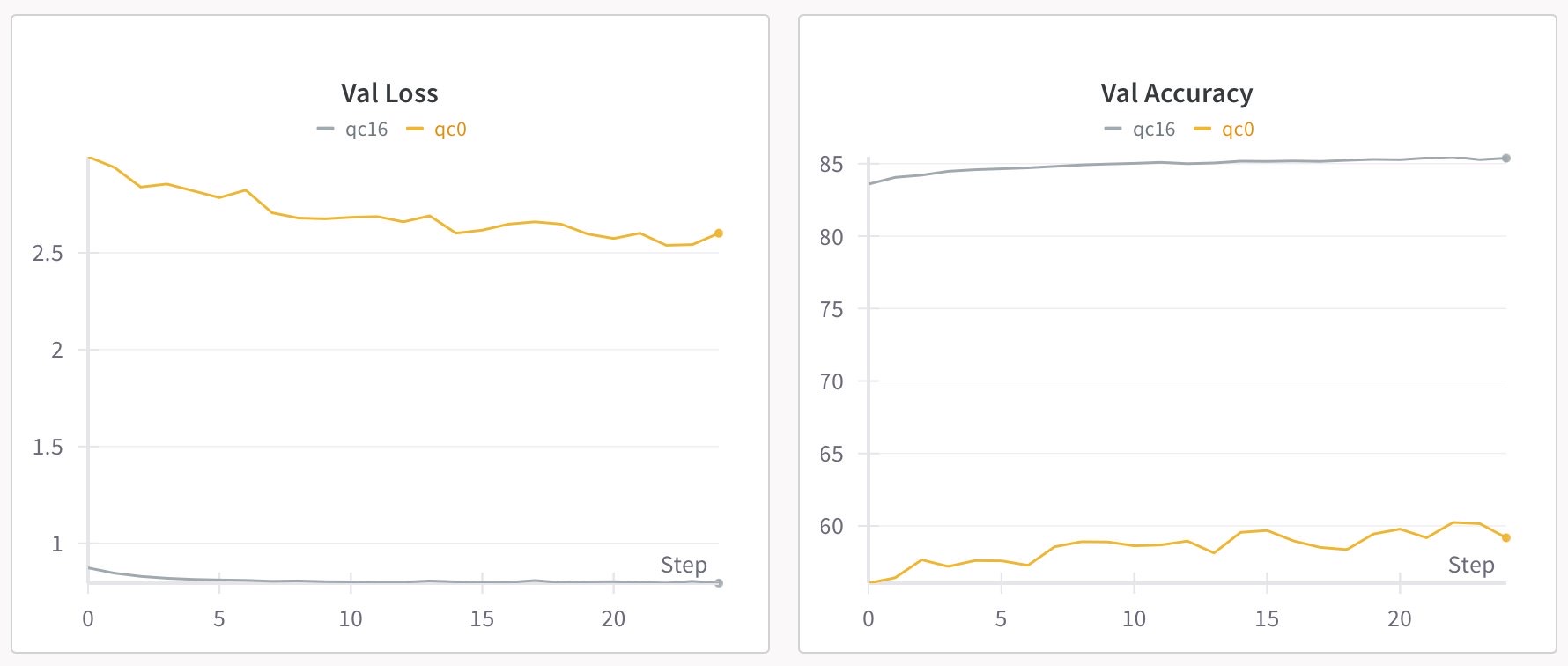}
        \caption{Qwen3-0.6B Chat. \textbf{Yellow}: training without synthetic pre-training (ShareGPT only).
        \textbf{Grey}: synthetic pre-training followed by fine-tuning.}
        \label{fig:qwen_chat_synth}
    \end{subfigure}
    \caption{Effect of synthetic data pre-training on inverse model optimization.
    Synthetic pre-training enables stable convergence and substantially higher validation accuracy, while training on limited real prompts alone fails to learn a meaningful inverse mapping.}
    \label{fig:synth_pretraining}
\end{figure*}

Figures~\ref{fig:qwen_base_synth} and~\ref{fig:qwen_chat_synth} illustrate the impact of synthetic data pre-training on inverse model performance for Qwen3 Base and Qwen3 Chat, respectively. In both cases, we compare two training regimes: (i) pre-training the inverse model on synthetically generated sequences followed by fine-tuning on 400 ShareGPT prompts, and (ii) training only on the same 400 ShareGPT prompts without synthetic pre-training. The results show a stark contrast between the two settings. Models trained without synthetic pre-training exhibit high validation loss and poor accuracy, failing to learn a meaningful inversion mapping from limited real prompts alone. 

In contrast, synthetic pre-training leads to rapid loss convergence and substantially higher validation accuracy, with fine-tuning providing only incremental refinement. This behavior is consistent across both base and chat variants, highlighting that large-scale synthetic probing is essential for learning the inverse generative structure of the forward LLM, and that small real-world prompt sets are insufficient on their own. These results underscore the central role of synthetic data in enabling stable and effective LLM inversion in a fully black-box setting.

\newpage
\section{Prompt-Response Reconstruction Histograms}\label{sec:pr}
\begin{figure*}[h]
    \centering
    \includegraphics[width=1\textwidth]{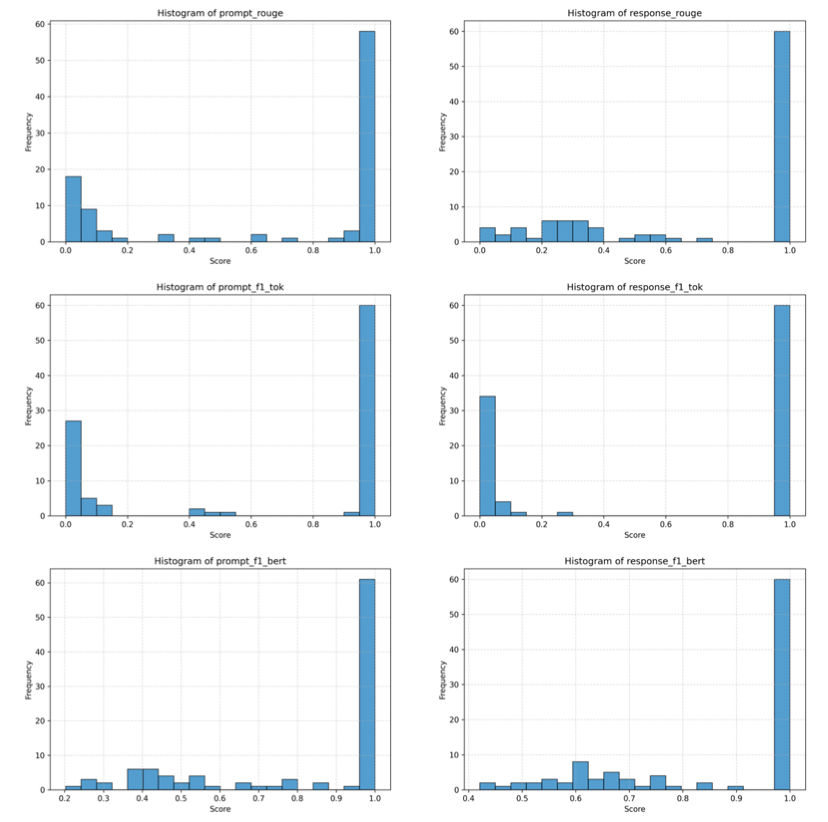}
    \caption{Distributions of prompt- and response-level reconstruction scores for
    \textbf{ROUGE-L}, \textbf{Token F1}, and \textbf{BERT F1}.
    Across all metrics, response-level distributions are more tightly concentrated than prompt-level ones,
    demonstrating that diverse reconstructed prompts consistently induce semantically similar responses.}
    \label{fig:hist_rouge_tokf1_bertf1}
\end{figure*}

\begin{figure*}[h]
    \centering
    \includegraphics[width=1\textwidth]{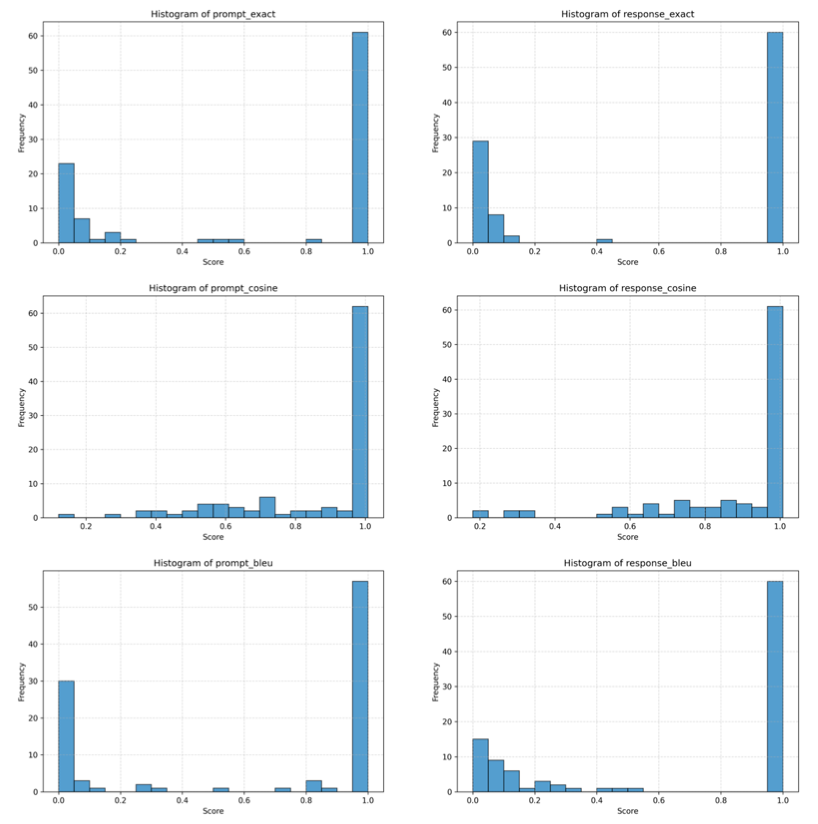}
    \caption{Distributions of prompt- and response-level reconstruction scores for
    \textbf{Exact Match}, \textbf{Cosine Similarity}, and \textbf{BLEU}.
    For each metric, the left histogram corresponds to prompt reconstruction and the right to response reconstruction.
    Prompt-level distributions exhibit higher variance, while response-level scores concentrate near high values,
    indicating functional equivalence despite prompt diversity.}
    \label{fig:hist_exact_cos_bleu}
\end{figure*}

To further analyze reconstruction behavior beyond aggregate scores, Figures~\ref{fig:hist_rouge_tokf1_bertf1} and~\ref{fig:hist_exact_cos_bleu} visualize the full distribution of reconstruction metrics using histograms, with prompt- and response-level scores shown side by side for each metric. Across all six metrics, prompt reconstruction exhibits a markedly bimodal distribution: a large mass near perfect scores (e.g., Exact Match, BLEU, ROUGE-L, Token F1 close to 1) corresponding to near-exact recoveries, alongside a smaller tail of lower scores reflecting alternative but valid prompt reconstructions. In contrast, the corresponding response-level histograms are significantly more concentrated toward high scores, with reduced variance across all metrics. This pattern indicates that even when prompt reconstruction is not exact at the token level, the reconstructed prompts still induce highly similar responses under the forward model.

\newpage
Notably, this effect is most pronounced for semantic metrics such as cosine similarity and BERT F1, where response distributions collapse tightly around high values despite broader dispersion at the prompt level. These histograms empirically reinforce the many-to-one nature of the forward generative mapping: diverse prompts uncovered by the inverse model often lie within the same response-equivalence class. Taken together, the histogram analysis complements the quantitative results in Tables~\ref{tab:prompt_recon} and~\ref{tab:response_recon}, demonstrating that our PTP-based inversion not only achieves strong average performance, but also reliably preserves functional response behavior across a wide spectrum of reconstructed prompts.

\newpage
\newpage
\section{Qualitative Results}\label{sec:qualitative_results}
We next look at some of the qualitative results obtained from our inverse LLM. These include both multiple prompt reconstructions for a given response as well as response reconstructions from recovered prompts. The evaluation is carried out not only on the unseen test prompts but also on random prompts, in order to illustrate the generality of the approach beyond the curated test set. These qualitative examples highlight that our method can recover exact prompts, generate semantically diverse alternatives, and still yield responses closely aligned with the original forward model outputs.

\subsection{Multiple Prompt Reconstruction} An important advantage of training an inverse LLM is the ability to generate \emph{multiple} plausible prompts for the same response. Since inference is performed via previous-token prediction, the inverse model can be sampled repeatedly to produce diverse reconstructions by varying decoding parameters such as top-$k$, top-$p$, and temperature. This goes beyond recovering a single deterministic prompt and instead enables exploration of the space of prompts that lead to semantically similar responses under the forward model.  

To illustrate, consider the above example. The original random prompt and response are shown, along with multiple reconstructed prompts sampled from the inverse model. This example highlights that while the model is able to recover the exact prompt, it can also generate diverse alternatives that preserve the semantic intent of the original prompt.

\begin{tcolorbox}[colback=gray!5!white,colframe=black!75!white,title=Example of Multiple Prompt Reconstruction]
\textbf{Prompt:} How to reach out to competitors to find their pricing strategies?  

\textbf{Response:} Reach out to competitors to gain insights into their pricing strategies and identify areas for improvement or innovation. Here are some steps you can follow:  

1. Identify Competitors: Determine the specific market or industry you want to compete in. Identify your closest competitors and other potential competitors.  

\dots  

7. Use the Insights: Use the information you gain from your competitors' responses to inform your own pricing strategy. Consider how their pricing strategies can be adapted to meet your own business needs.  

\textbf{Reconstructed Prompts:}  

1. What do you need to know about their pricing strategies to invest in market research, competitive analysis, and competitive planning?  

2. What are their strengths and weaknesses in their pricing strategies? What are the key challenges that my companies face in their pricing strategies?  

3. How to reach out to competitors to find their pricing strategies?  

4. How do you identify the most effective marketing strategies to attract potential clients? Choose from a wide range of competitors and provide insights into their pricing strategies? 

5. Can you provide me with a list of pricing strategies for businesses that compare to its competitors? Also, are you looking to target competitors and their pricing strategies for their product/service?  

6. What do we need to know about the potential opportunities for marketers to reach out to competitors to find their pricing strategies?  

7. What tactics can a company looking to reach out to competitors in the market use to find their pricing strategy?  
\end{tcolorbox}

\newpage

\subsection{Prompt Reconstruction on ShareGPT}
To further evaluate our method beyond synthetic sequences, we test the inverse model on real-world prompts drawn from the ShareGPT dataset. In this setting, prompts are selected from ShareGPT, their responses are generated by the forward model, and the inverse model is tasked with reconstructing the original prompts from the responses alone.  We present examples below for both the Qwen Base (QB) and Qwen Chat (QC) variants. These qualitative cases illustrate that while the reconstructed prompts may vary in surface form, they often remain semantically faithful to the original intent.

\begin{tcolorbox}[colback=gray!5!white,colframe=black!75!white,title=Qwen Base on ShareGPT (Unseen Test Prompt)]
\textbf{Inverse Model Input:} The expiration effect in equities refers to the phenomenon where the price of an option moves in a way that is not solely driven by the intrinsic value of the option but also by the time value of the option. This effect is particularly pronounced in options with shorter expiration dates, such as call options. Here’s a fundamental reason behind this effect and how to exploit it in a trading strategy:  
Fundamental Reason Behind the Expiration Effect

1. **Time Value**: The time value of an option is the difference between its premium and its intrinsic value. As the expiration date approaches, the time value of the option decreases because the probability of the option expiring in-the-money diminishes.\dots

\textbf{Inverse Model Output (Reconstructed Prompt):}  
Give me a brief explanation of the fundamental reason behind the expiration effect in equities and how to exploit this information in trading.  

\textbf{Actual Prompt:}  
Fundamental reason behind option expiration effect in equities and how to exploit this information in creating a simple trading strategy with code.  
\end{tcolorbox}

\begin{tcolorbox}[colback=gray!5!white,colframe=black!75!white,title=Qwen Chat on ShareGPT (Unseen Test Prompt)]
\textbf{Inverse Model Input:} Let me know if there are any other steps I could take.  
Yes, creating an advertising campaign on Pinterest can be an effective strategy. Here are some examples of steps you could take:  
1. Define your campaign goals: Set a specific objective such as increasing brand awareness or driving sales.  
2. Create visually appealing content: Use high-quality images and videos that reflect Semilac's values and products.  
\dots
6. Collaborate with influencers: Partner with influencers who have a strong presence on Pinterest to amplify your brand's reach.  
7. Run ad campaigns  

\textbf{Inverse Model Output (Reconstructed Prompt):}  
Give me examples of steps I could take to create an advertising campaign on Pinterest for Semilac brand. How can I use this information to make the campaign effective?  

\textbf{Actual Prompt:}  
Give me examples of steps I could take to create an advertising campaign on Pinterest for Semilac brand.  
\end{tcolorbox}

\newpage
\subsection{Prompt--Response Reconstruction on Open Questions}
Beyond prompt-only reconstruction, we also evaluate whether the reconstructed prompts, when passed back into the forward model, yield responses that are comparable to the originals. This experiment effectively closes the loop: it tests whether the inverse model can recover prompts that not only resemble the originals but also elicit semantically consistent downstream generations. This criterion is stricter and more practical than prompt reconstruction alone, since it validates that the reconstructed prompt preserves intent in interaction with the forward model.  

\begin{tcolorbox}[colback=gray!5!white,colframe=black!75!white,title=Prompt--Response Reconstruction Example]
\textbf{Prompt:} Is it appropriate to coordinate pricing with competitors that sell the same product and service?  

\textbf{Response:} Coordinating pricing with competitors that sell the same product or service can be appropriate in certain contexts, but it depends on the company's goals, market conditions, and strategic priorities. Here's a breakdown:  
1. \textbf{Strategic Considerations}: Competitive Edge: If the company aims to maintain a competitive advantage, pricing coordination with similar competitors can help differentiate offerings. ...
7. \textbf{Potential Risks}: Loss of Competitive Advantage: Overcoordination may lead to undercutting competitors… Market Confusion: If competitors use similar strategies, customers might become confused… 

\textbf{Reconstructed Prompt:} What are some ways to reduce the difficulty level in the market for a product that is double-crossing a new venture with a supplier? Additionally, what are the key factors that make the company's operations a good fit for the market in the context of the above pricing strategy?  

\textbf{Reconstructed Response:}  

1. \textbf{Reducing Market Difficulty for a Supplier/Double-Crossing Venture}: Diversify suppliers; enhance risk management…

2. \textbf{Key Factors for a Good Fit in Operations}: Alignment with company goals; efficient resource allocation…

7. \textbf{Key Factors for a Good Fit in the Market Context of Pricing Strategy}: Understanding customer needs; analyzing competitor pricing; ensuring pricing reflects costs and margins; adapting to market trends. 

Each factor contributes to a balanced and sustainable business model in both supplier and market contexts.  
\end{tcolorbox}

In the example shown below, the reconstructed prompt contains multiple questions, reflecting how the inverse model leverages contextual cues present in the original response to generate plausible alternatives. Despite this structural variation, when the reconstructed prompt is fed back into the forward model, the resulting response aligns closely with the original, demonstrating semantic consistency. Notably, the reconstructed response includes the phrase ``double-crossing a new venture with a supplier,'' which arises directly from the model’s attempt to interpret the original question about coordinating pricing with competitors. This illustrates how the inverse model not only reconstructs the surface form of the prompt but also re-injects contextual elements from the response into the reconstructed prompt.

\newpage

\subsection{Transferability}
In transferability we are interested in evaluating whether an inverse model trained on one forward LLM can generalize to responses generated by a different model. To evaluate this, we test the inverse model trained on Qwen-3 0.6B Chat (QC) against responses produced by GPT-4o. Specifically, we provide a random prompt to GPT-4o, obtain its response, reverse the responses at the token level, and then use the inverse model to reconstruct the prompt. The reconstructed prompt is then fed back into GPT-4o to generate a reconstructed response. This setup tests whether the inverse model preserves transferability across model families and can still yield semantically meaningful prompt--response reconstructions.

\begin{tcolorbox}[title=Transferability Example,colback=white,colframe=black]
\textbf{Prompt:} How to reach out to competitors to find their pricing strategies? \\[6pt]

\textbf{Response:} Reaching out to competitors to understand their pricing strategies needs to be done ethically and legally. Directly asking competitors for their pricing details or strategies could lead to anti-competitive behavior, which might violate laws like antitrust regulations in many jurisdictions. Here are some ethical and appropriate ways to gain insights into competitor pricing strategies:  
\begin{itemize}
    \item \textbf{Conduct Market Research:} Monitor their public pricing through websites, catalogs, or brochures.  
    \item \textbf{Leverage Mystery Shopping:} Pose as a customer to learn about pricing directly from sales representatives, while remaining within ethical boundaries.  
    \item \textbf{Study Their Marketing Strategies:} Analyze advertising campaigns to infer pricing focus, such as low prices, premium value, or bundled packages.  
\end{itemize}
\textit{Important Ethical and Legal Considerations:} Do not engage in price fixing or collusion, as these are illegal in most countries. Instead, focus on aligning pricing with value propositions, cost structure, and customer willingness to pay. \\[8pt]

\textbf{Reconstructed Prompt:} Can you provide information on the current and future trends in their pricing strategies? Also, can you suggest ways to increase their market share in the market? \\[6pt]

\textbf{Reconstructed Response:} Certainly! Below is an overview of current and future trends in pricing strategies, along with strategies to increase market share.  
\begin{itemize}
    \item \textbf{Dynamic Pricing:} Adjust prices in real time based on demand, competitor pricing, or customer profiles.  
    \item \textbf{Value-Based Pricing:} Align prices with customer-perceived value to maximize revenue.  
    \item \textbf{Strategies to Increase Market Share:} Enhance product differentiation, strengthen brand positioning, and implement targeted promotions to attract customers.  
\end{itemize}
By staying attuned to industry trends and applying these strategies, companies can remain competitive and expand their market presence.  
\end{tcolorbox}

This example illustrates that even when the forward responses originate from a model different from the one used to train the inverse model, the inverse LLM still reconstructs semantically aligned prompts and responses. While the reconstructed prompt is not identical to the original, it captures the underlying context and intent, demonstrating transferability across model families.


\newpage

\end{document}